# Bioformer: an efficient transformer language model for biomedical text mining

Li Fang[1,2,†], Qingyu Chen[3,†], Chih-Hsuan Wei[3], Zhiyong Lu[3], and Kai Wang[2,4*]

1. Department of Genetics and Biomedical Informatics, Zhongshan School of Medicine, Sun Yat-sen University, Guangzhou, 510080, China

2. Raymond G. Perelman Center for Cellular and Molecular Therapeutics, Children's Hospital of Philadelphia, Philadelphia, PA 19104, USA

3. National Center for Biotechnology Information (NCBI), National Library of Medicine (NLM), National Institutes of Health (NIH), Bethesda, MD 20892, USA

4. Department of Pathology and Laboratory Medicine, University of Pennsylvania Perelman School of Medicine, Philadelphia, PA 19104, USA

† These authors contributed equally: Li Fang and Qingyu Chen

* Correspondence: wangk@chop.edu (K.W.)

## Abstract

Pretrained language models such as Bidirectional Encoder Representations from Transformers (BERT) have achieved state-of-the-art performance in natural language processing (NLP) tasks. Recently, BERT has been adapted to the biomedical domain. Despite the effectiveness, these models have hundreds of millions of parameters and are computationally expensive when applied to large-scale NLP applications. We hypothesized that the number of parameters of the original BERT can be dramatically reduced with minor impact on performance. In this study, we present Bioformer, a compact BERT model for biomedical text mining. We pretrained two Bioformer models (named *Bioformer$_{8L}$* and *Bioformer$_{16L}$*) which reduced the model size by 60% compared to BERT$_{Base}$. Bioformer uses a biomedical vocabulary and was pre-trained from scratch on PubMed abstracts and PubMed Central full-text articles. We thoroughly evaluated the performance of Bioformer as well as existing biomedical BERT models including *BioBERT* and *PubMedBERT* on 15 benchmark datasets of four different biomedical NLP tasks: named entity recognition, relation extraction, question answering and document classification. The results show that with 60% fewer parameters, *Bioformer$_{16L}$* is only 0.1% less accurate than *PubMedBERT* while *Bioformer$_{8L}$* is 0.9% less accurate than *PubMedBERT*. Both *Bioformer$_{16L}$* and *Bioformer$_{8L}$* outperformed *BioBERT$_{Base}$-v1.1*. In addition, *Bioformer$_{16L}$* and *Bioformer$_{8L}$* are 2-3 fold as fast as PubMedBERT/*BioBERT$_{Base}$-v1.1*. Bioformer has been successfully deployed to PubTator Central providing gene annotations over 35 million PubMed abstracts and 5 million PubMed Central full-text articles. We make Bioformer publicly available via https://github.com/WGLab/bioformer, including pre-trained models, datasets, and instructions for downstream use.

## Introduction

The volume of published biomedical literature is increasing rapidly over the past few years. For instance, PubMed has more than 35 million articles, and this number is growing by five thousand per day. In addition, there are 8.7 million freefull-text articles available in PubMed Central (PMC) and preprint servers, such as bioRxiv and medRxiv. Such rapid growth challenges knowledge discovery and literature curation. Biomedical Natural Language Processing (BioNLP) has been applied to help decrease such burdens. BioNLP language models–capturing semantic representations over biomedical literature – are the foundations for BioNLP method development and downstream applications. Early BioNLP language models include biomedical word embeddings [1, 2], entity embeddings [3, 4], and sentence embeddings [5] have shown effectiveness in a range of BioNLP applications such as biomedical named entity recognition, relation extraction, and information retrieval [6, 7].

Recently, pretrained transformer language models, such as Bidirectional Encoder Representations from Transformers (BERT) [8], have led to impressive performance gains over word/sentence embeddings, convolutional neural networks (CNNs) and recurrent neural networks (RNNs). BERT uses the transformer architecture [9] and it is pretrained by self-supervised learning. It has been adapted to the biomedical domain: pretrained BERT models on biomedical literature or clinical text, such as BioBERT [10], BlueBERT [11], PubMedBERT [12] and BioClinicalBERT(https://huggingface.co/emilyalsentzer/Bio_ClinicalBERT)[13], have been released to the public. While these BERT models could improve the effectiveness of downstream BioNLP studies, they have hundreds of millions of parameters that are computationally expensive. For instance, existing studies have shown BioNLP BERT models are 50 times slower than biomedical sentence embeddings for retrieving relevant sentences via biomedical literature [14]. In addition, a full BioNLP pipeline usually includes multiple subtasks and therefore requires running BERT models multiple times. For example, a pipeline to extract disease-casual genes from biomedical literature needs first to identify named entities of genes and diseases. Next, entity normalization needs to be performed to link the gene and disease mentions to unique identifiers. After that, a relation extraction model is needed to classify gene-disease relations. This pipeline runs BERT models five times if each subtask uses its own fine-tuned model, not to mention it may need to be applied to millions of biomedical literatures. Therefore, it is the bottleneck for BioNLP studies to apply transformer-related language models and deploy them in real-world applications for biomedical researchers and healthcare professionals.

The existing BioNLP transformer models directly apply the $BERT_{Base}$ or $BERT_{Large}$ architecture with 110-340 million parameters. To date, there is no BioNLP transformer on improving efficiency. To speed up large-scale or real-time biomedical text mining tasks, we aim to train a compact model with faster speed while maintaining high accuracy. In the general domain, recent studies[15] showed that simply pretraining longer and over more data significantly improved the performance. Previous studies [16] also showed that given a fixed model size, model depth (number of layers) and width (hidden embedding size) have a large effect on performance. Therefore, we hypothesize that a well pretrained compact model with optimal model depth and width could achieve faster speed with minimal loss in accuracy.

In this study, we present Bioformer, a compact BERT model for biomedical text mining. Bioformer has two variants: *Bioformer$_{8L}$* and *Bioformer$_{16L}$* with different model depth and width. Bioformer uses a biomedical vocabulary and is pretrained from scratch on 33 million PubMed abstracts and 1 million PubMed Central (PMC) full-text articles. We compared its efficiency with existing biomedical BERT models such as *BioBERT* and PubMedBERT. We thoroughly evaluated the effectiveness of Bioformer over 15 datasets from four primary BioNLP tasks: named entity recognition, relation extraction, question answering and document classification. Our results show that with 40% of the parameters, *Bioformer$_{16L}$* is only 0.1% less accurate than *PubMedBERT* and its overall performance is even better than *BioBERT$_{Base}$*-v1.1. In addition, *Bioformer* models are 2-3-fold as fast as *PubMedBERT*/*BioBERT$_{Base}$*-v1.1. Bioformer has been successfully deployed to PubTator Central [17], providing automatic annotations over 35 million PubMed abstracts and 5 million PubMed Central full-text articles.

## Methods

As shown in Table 1, Existing BioNLP BERT models directly applies $BERT_{Base}$ with the total number of parameters of 110M ($L$ = 12, $H$ = 768, $A$ = 12) where $L$, $H$, and $A$ stand for the number of layers, the hidden embedding size, and the number of attention heads, respectively. We hypothesize that the hundred million of parameters in the BERT architecture are not equally effective and some might be redundant. Instead, we empirically investigated compact architectures by dramatically reducing the number of layers, the hidden embedding size, and the number of attention heads. Specifically, we pretrained two compact models detailed in Table 1: *Bioformer$_{8L}$* with the total number of parameters of 43M ($L$ = 8, $H$ = 512, $A$ = 8) and *Bioformer$_{16L}$* ($L$ = 16, $H$ = 1024, $A$ = 6). They have less than 40% of parameters compared to that of the original BERT$_{Base}$. *Bioformer$_{8L}$* and *Bioformer$_{16L}$* have almost the same number of parameters; in contrast, *Bioformer$_{16L}$* has more layers and a smaller embedding size (i.e., deeper and thinner).

## Vocabulary of Bioformer

Bioformer uses a cased WordPiece (a subword segmentation algorithm [18]) vocabulary trained from all PubMed abstracts (33 million, as of Feb 1, 2021) and one million subsampled PMC full-text articles. We subsampled one million PMC full-text articles so that total size of PubMed abstracts and PMC full-text articles are approximately the same. In the training process, the vocabulary was initialized with individual characters in the corpus; then, the most frequent subwords in the corpus were iteratively added to the vocabulary. We trained Bioformer's vocabulary from the Unicode text of the two resources to mitigate the out-of-vocabulary issue and include special symbols (e.g., male and female symbols) in biomedical literature.

## Pretraining Bioformer

The pretraining details for Bioformer is summarized in Table 1. The workflow for pretraining Bioformer is shown in Figure 1. Bioformer was pretrained from scratch on the same corpus as the vocabulary in 2.1. The original BERT model has two pretraining objectives: masked language model (MLM) and next sentence prediction (NSP). The MLM objective is for predicting the masked tokens (subwords). The NSP objective is to predict whether two input sentences are next to each other in the original text. For the MLM objective, we used whole-word masking with a masking rate of 15%. Whole-word masking requires that the whole word (i.e., all the subwords) be masked if one of its subwords is chosen (https://github.com/google-research/bert). This will force the model to recover the whole word using context information instead of recovering a subword, which can be inferred from the unmasked subwords of the same word. The random masking process was duplicated 20 times so that each sequence was masked in 20 different ways. There are debates on whether the NSP objective could improve the performance on downstream tasks[15]. We include it in our pretraining experiment in case the prediction of the next sentence is needed by downstream tasks (e.g., for zero-shot learning [19]). Sentence segmentation of all training text was performed using SciSpacy [20]. Pretraining of Bioformer was performed on a single Cloud TPU device (TPUv2, 8 cores, 8GB memory per core). We pretrained Bioformer (both models) for 2 million steps with a batch size of 256. It took about 8.7 days to finish the pretraining of *Bioformer$_{8L}$* and 11.3 days to finish the pretraining of *Bioformer$_{16L}$*.

## Evaluation of training and inference speed

We evaluated the training (for fine-tuning) and inference speed based on a sequence classification task. The sequence classification task adds a linear layer on top of the [CLS] token for binary classification. Therefore, it has minimal changes to the BERT backbone. We evaluated the speed of four biomedical BERT models (*BioBERT$_{Base}$*, *PubMedBERT, Bioformer$_{8L}$, Bioformer$_{16L}$*) as well as one compact general domain BERT model: DistilBERT. DistilBERT was a distilled version of BERT developed by Hugging Face[15]. It has the same hidden embedding size as *BERT$_{Base}$* but with fewer number of layers ($L$=6). We used the 'run_glue.py' script in the Transformers library[21] to perform the benchmark. Training speed was assessed on an NVIDIA Tesla P100 GPU with 16GB memory. Inference speed was assessed on an Intel Xeon CPU. The max sequence length was set to 512. For training on GPU, the batch size was set to the maximal possible value for each model that does not cause out-of-memory error.

# Evaluation of performance on biomedical NLP tasks

Overall, we evaluated the performance of five biomedical BERT models: *Bioformer$_{8L}$*, *Bioformer$_{16L}$*, *BioBERT$_{Base-v1.1}$*, *PubMedBERT$_{Abs}$* (pretrained on PubMed Abstracts only), and *PubMedBERT$_{AbsFull}$* (pretrained on PubMed Abstracts and PMC full-text articles). To ensure fairness, all the five models were evaluated using the same setting. We evaluated the performance on four main biomedical NLP applications consisting of 15 datasets: named entity recognition (eight datasets), relation extraction (four datasets), question answering (one dataset), and document classification (two datasets). The details of the datasets and evaluation metrics are summarized in Table 2. We used consistent preprocessing and training approaches as the existing studies summarized below.

## Named Entity Recognition (NER)

Named Entity Recognition identifies named entities mentioned in unstructured text, such as gene and disease names. It is commonly formulated as a token classification problem: for a given token in the input text, the model needs to determine whether it belongs to a specific entity type. We adopted the same preprocessing pipeline for NER applications in the existing studies [10, 12]. The context-ware representation of each token from the last layer of a transformer model is used to classify the token's category.

## Relation Extraction (RE)

Relation extraction classifies the relationships between named entities in the given text (e.g., protein-protein interactions). It can be formulated as a sequence classification problem: given a passage, it classifies whether it mentions a specific relation. We followed the preprocessing method used by BioBERT[10] where entity names are replaced by pre-defined tags (e.g., gene and disease names are replaced (@GENE$ and @DISEASE$, respectively). The representation of the [CLS] token in the last layer is used to classify the relations.

## Question Answering (QA)

Question answering is a reading comprehension task that extracts the answer to a question from a given text. The Stanford Question Answering Dataset (SQuAD)[22] is a large-scale QA dataset in the general domain. BioASQ[23] factoid datasets are a series of QA datasets in the biomedical domain. Given a question and a passage containing the answer, the task is to predict the span of the answer. We followed the same fine-tuning procedure as the original BERT[8] for QA tasks. The input question and passage are presented as a sentence pair where the question is the first sentence and the passage is the second sentence. Token-level probabilities for the start/end location of the answer span are computed using a single layer. For BioASQ, we used the preprocessed dataset (BioASQ-7b) provided by the developers of BioBERT[10] where about 30% unanswerable questions had been removed from the dataset. We used the same evaluation metrics from BioASQ: strict accuracy, lenient accuracy and mean reciprocal rank. Previous studies[10, 24] showed that pretraining on the SQuAD dataset (fine-tuning on the SQuAD before fine-tuning on BioASQ) improved the performance on the BioASQ datasets. Therefore, we showed the evaluation results with or without pretraining on SQuAD.

### Document Classification (DC)

Document classification categorizes an input document into one or more categories. In contrast to relation extraction which focuses on the relationships between entities often at passage level, document classification is often at abstract or full-text level. It has been extensively used for biomedical document triage (e.g., classify whether a PubMed article is relevant for manual curation [25] and topic assignment (e.g., assign COVID-19-related topics such as Treatment and Diagnosis to a relevant article [26]). Same as the relation extraction task, the representation of the [CLS] token in the last layer is used to classify the document.

### Hyperparameter tuning and result reporting

For each task, we selected the max sequence length based on previous studies and the values were listed in Table 2. Consistent with the existing studies[12], we select the best batch size among 8, 16, and 32, and learning rate among 1e-5, 3e-5, and 5e-5 on the development set. For each combination, we trained the model for up to 20 epochs, selected the checkpoint achieving the best performance over the development set, and evaluated it on the testing set. To ensure a fair comparison, we used the same preprocessing and hyperparameter tuning methods for all the five models. We repeated each fine-tune experiments for five times using different random seeds and reported the average performance of the evaluation metrics in Table 2.

## Results

### Pretraining Bioformer and hyperparameter selection

Model depth (number of transformer layers, $L$) and model width (hidden embedding size, $H$) are key hyperparameters that impact the performance. Turc et al.[16] released 24 pretrained miniature BERT models in the general domain. To investigate the best model depth and model width for a compact BERT model, we examined the effect of model depth and width on performances of downstream tasks. We compared decreasing depth versus decreasing width. We benchmarked the performances on general domain NER and QA tasks (**Figure 2**). NER is a relatively simple task and QA is a more complicated task. The performance drop for QA is more significant than that for NER. We also observed that the performance drop is higher when reducing depth ($L$) compared to reducing width ($H$). This suggest that a deep-and-narrow model might perform better than a shallow-and-wide model when the model sizes are approximately the same. Therefore, we pretrained two Bioformer models with approximately the same model size: *Bioformer$_{8L}$* and *Bioformer$_{16L}$* (hyperparameters shown in Table 1). Compared to a BERT$_{Base}$ model, *Bioformer$_{8L}$* reduced both model width (from 768 to 512) and depth (from 12 to 8). *Bioformer$_{16L}$* reduced model width more significantly (from 768 to 384), but increased model depth to 16. We hypothesized that *Bioformer$_{16L}$* might perform better than *Bioformer$_8$* on more complicated task such as QA. The procedure for pretraining *Bioformer$_{8L}$* and *Bioformer$_{16L}$* is described in the method section.

### Model structure and speed

We compared the speed of Bioformer and with *BERT$_{Base}$*/*BioBERT$_{Base}$*, *PubMedBERT* and *DistilBERT*. (Figure 3). *BioBERT$_{Base}$* used a continue pretraining strategy and was initialized with model weights from the original *BERT$_{Base}$*. Therefore, *BioBERT$_{Base}$* and *BERT$_{Base}$* have the same speed. *PubMedBERT* have approximately the same structure as *BERT$_{Base}$*, except that it has its own vocabulary with a slightly different vocabulary size. *DistilBERT* was a compact BERT model developed by Hugging Face[27]. *DistilBERT* has the same hidden embedding size as *BERT$_{Base}$* but with fewer number of layers (*L*=6). As a result, the model size was reduced by 40%. *Bioformer$_{8L}$* and *Bioformer$_{16L}$* have approximately the same model size (43M and 42M, **Figure 3C**), but *Bioformer$_{16L}$* is a deeper model with smaller hidden embedding size. We benchmarked the training and inference speed of these models on sequence classification task which added a single linear layer on top of the [CLS] token. The max sequence length was 512. The benchmarking results are shown in Figure 3. *PubMedBERT* and *BioBERT$_{Base}$* has approximately the same speed. *Bioformer$_{16L}$* is twice as fast as *BioBERT$_{Base}$* for training and the inference speed is 2.2-fold of *BioBERT$_{Base}$*. The training and inference speeds of *Bioformer$_{8L}$* are 2.8-fold and 3.0-fold of BioBERT$_{Base}$, respectively. The speed of *Bioformer$_{16L}$* and *DistilBERT* are approximately the same while *Bioformer$_{8L}$* is faster than *Bioformer$_{16L}$* and *DistilBERT*.

## Performance on downstream biomedical NLP tasks

Table 3 shows the performance on four downstream biomedical NLP tasks: named entity recognition, relation extraction, question answering and document classification. Overall, with only 40% of the parameters of the original *BERT$_{Base}$*, *Bioformer$_{16L}$* and *Bioformer$_{8L}$* retain high accuracy across the 15 datasets of the four tasks. *PubMedBERT$_{Abs}$* achieved the highest overall average performance score among all models, followed by *Bioformer$_{16L}$*. *Bioformer$_{16L}$* retains 99.92% (82.71/82.77) performance score of *PubMedBERT$_{Abs}$* and its average performance is better than *PubMedBERT$_{Abs}$*, *BioBERT$_{Base-v1.1}$* and *Bioformer$_{8L}$*. The overall performance of *Bioformer$_{8L}$* is a little higher than *BioBERT$_{Base-v1.1}$* (average score 82.07 versus 82.02). *Bioformer$_{16L}$* has better performance than *Bioformer$_{8L}$* in most (12 out of 15) datasets, which is consistent with the finding that depth is more important than width (Figure 2 and also reported by previous studies [16, 28]). We also observed that *Bioformer$_{16L}$* performed best in the QA task, outperformed *Bioformer$_{8L}$* and other models by a large margin.

Of note, *Bioformer$_{8L}$* participated the BioCreative VII LitCovid multi-label topic classification challenge and achieved the overall best performance among all teams [29]. The details of the challenge are described elsewhere [29, 30]. In this study, we showed the performance of the five models on the BioCreative VII LitCovid challenge development set in Table 3 (as a document classification task). The performance scores in Table 3 are different from the BioCreative VII LitCovid challenge paper[30] because Table 3 shows the average scores of five random runs while the BioCreative VII LitCovid challenge paper presented the score of the submitted run. *Bioformer$_{16L}$* did not participate this challenge because *Bioformer$_{16L}$* was pretrained after the challenge was closed. We added the performance score of *Bioformer$_{16L}$* in Table 3.

## Incorporating Bioformer into PubTator for annotating genes at the full PubMed and PMC scale

We further demonstrate a real-world application by incorporating Bioformer into PubTator. PubTator [17] – a web-based biomedical text mining platform publicly available for over a decade – provides high-quality automatic entity annotations and prioritizes target entities or relevant articles for downstream research. PubTator provides six types of biomedical concepts (genes, chemicals, diseases, species, variants, and cell lines) for the entire PubMed and PMC articles (to date, there are 35+ million PubMed abstracts and 5 million+ PMC full texts).

Previously, PubTator employed the conditional random fields (CRF) for recognizing genes[17]. This approach took about 100 hours to annotate genes over the entire PubMed and PMC articles in the production environment consisting of a computer cluster of 300 CPUs. Given the advances in transformers, we implemented NER models using the existing BioNLP transformers as the backbone and examined the efficiency in the PubTator production environment. It took more than 350 hours for the existing BioNLP transformers to annotate genes. As PubTator annotates seven types of named entities, it is not practical to extend the annotations to all the seven types at the full PubMed and PMC scale.

We used *Bioformer$_{8L}$* instead. Under the same resources, the processing time significantly reduced to 150 hours, whereas the accuracy difference is within 1% [31]. The processing time is competitive with that of the previous CRF method. This demonstrates a successful use case of incorporating Bioformer into PubTator and bringing dramatic efficiency to large-scale data processing.

## Discussion

In this study, we introduced Bioformer, a compact pretrained language model for biomedical text mining. Bioformer has two variants (*Bioformer$_{8L}$ and Bioformer$_{16L}$*) with approximately the same model size but different model widths and depths. We evaluated the effectiveness of Bioformer on 15 datasets of four biomedical NLP tasks. With 40% of the parameters, both variants retain more than 99% performance score of *PubMedBERT*, the current state-of-the-art model. The efficiency evaluation results showed that *Bioformer$_{16L}$* and *Bioformer$_{8L}$* and are 2X and 3X as fast as *PubMedBERT*, respectively. Bioformer has been deployed to PubTator Central, providing automatic annotations over 35 million PubMed abstracts and 5 million PubMed Central full-text articles.

The effectiveness evaluation results do not suggest significant difference for the five models in terms of accuracy. In contrast, the efficiency evaluation results show that Bioformer significantly improved the training and inference speed by 2-3 folds. These collectively suggest that Bioformer can be employed in applications where speed is critical (e.g. large-scale data analysis or real-time applications) while maintaining high accuracy. Bioformer reduced the model size by 60% which indicates that it can be fit into devices with smaller memory, or be trained with larger batch size given a fixed GPU/TPU memory. Recent studies suggest that pretraining with larger batch size may improve the performance and the state-of-the-art models such as RoBERTa, XLNet, and PubMedBERT were pretrained with a batch size of 8192 [12, 15, 31].

In addition, we also comparatively analyzed *Bioformer$_{8L}$* and *Bioformer$_{16L}$* and found a trade-off between effectiveness and efficiency. Recall that both models have a very similar number of parameters but different architectures. *Bioformer$_{8L}$* has eight transformer layers and the hidden embedding size is 512. *Bioformer$_{16L}$* has 16 transformer layers and the hidden embedding size is 384. The effectiveness evaluation shows that *Bioformer$_{16L}$* consistently had better performance than *Bioformer$_{8L}$* on most (12/15) datasets, which suggests that deeper and thinner transformer models may have higher accuracy. As a trade-off, *Bioformer$_{16L}$* is less efficient than *Bioformer$_{8L}$*, despite that *Bioformer$_{8L}$* has a little more number of parameters (43M versus 42M). The reason might be that *Bioformer$_{8L}$* has a larger hidden embedding size and large dense matrix multiplications can be well parallelized. *Bioformer$_{16L}$* has more layers and matrix multiplications in different layers cannot start simultaneously. The calculations of the next layer need the results from the previous layer.

*Bioformer$_{8L}$* retains 99.6% F1 score of *Bioformer$_{16L}$* in NER tasks. Considering the speed advantage, we suggest using *Bioformer$_{8L}$* for NER tasks if accuracy is not critical. However, *Bioformer$_{8L}$* only retains 95.6% of the average performance score of *Bioformer$_{16L}$* in question-answering tasks. Therefore, we suggest using *Bioformer$_{16L}$* for question-answering tasks considering its significant advantage. For relation extraction tasks, *Bioformer$_{8L}$* retains 98.4% of the average performance score of *Bioformer$_{16L}$*, users can choose the model based on their preference on speed/accuracy. We hope that Bioformer is a compelling option for large-scale text mining or real-time NLP applications.

## Acknowledgment

This study is supported by the Google TPU Research Cloud (TRC) program (L.F.), the Intramural Research Program of the National Library of Medicine (NLM), National Institutes of Health (NIH), and NIH/NLM grants LM012895 (K.W.) and 1K99LM014024-01 (Q.C.).

## Data and Code Availability

Pretrained model weights of *Bioformer$_{8L}$* and *Bioformer$_{16L}$* are publicly available on HuggingFace (https://huggingface.co/bioformers/bioformer-8L, https://huggingface.co/bioformers/bioformer-16L). The code and instructions for downstream use are available on GitHub (https://github.com/WGLab/bioformer).

# Figures

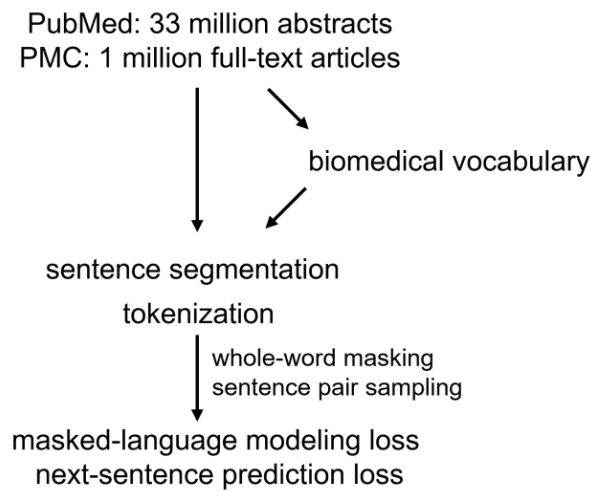

**Figure 1. Workflow for pretraining Bioformer.** We first trained a WordPiece vocabulary using all PubMed abstracts (33 million, as of Feb 1, 2021) and one million subsampled PMC full-text articles. The PMC full-text articles were subsampled to one million such that the total size of PubMed abstracts and PMC full-text articles are approximately the same. Bioformer models were pretrained from scratch using the same text for training the vocabulary. Same as the original BERT model, there were two pretraining objectives: masked language modeling (MLM) and next sentence prediction (NSP). For the MLM objective, we used whole-word masking with a masking rate of 15%. The random masking process was duplicated 20 times so that each sequence was randomly masked in 20 different ways. Sentence segmentation of all training text was performed using SciSpacy [20]. Pretraining of Bioformer was performed on a single Cloud TPU device (TPUv2, 8 cores, 8GB memory per core) with a max sequence length of 512 and a batch size of 256.

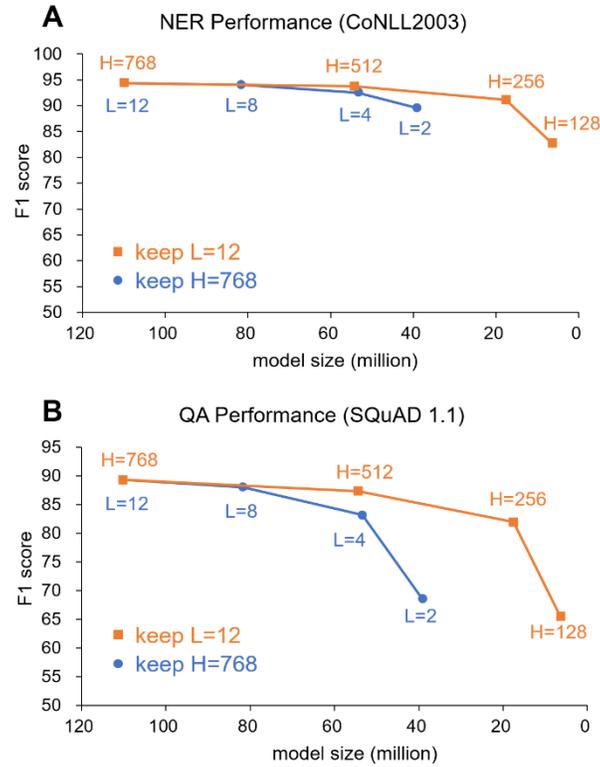

**Figure 2. Effect of model width (hidden embedding size, *H*) and depth (number of transformer layers, *L*) on performances of downstream tasks**. Two model compression methods were assessed: decreasing depth (keep width) or decreasing width (keep depth). We benchmarked the performances on NER (**A**) and question answering (**B**). The performance drop is higher when reducing depth (*L*) compared to reducing width (*H*). On the other words, a deep-and-narrow model performs better than a shallow-and-wide model when the model sizes are approximately the same. The compressed models used in this figure were general domain BERT models released by a previous study [16]. The datasets for benchmarking NER and QA performance are CoNLL2003 and SQuAD1.1, respectively.

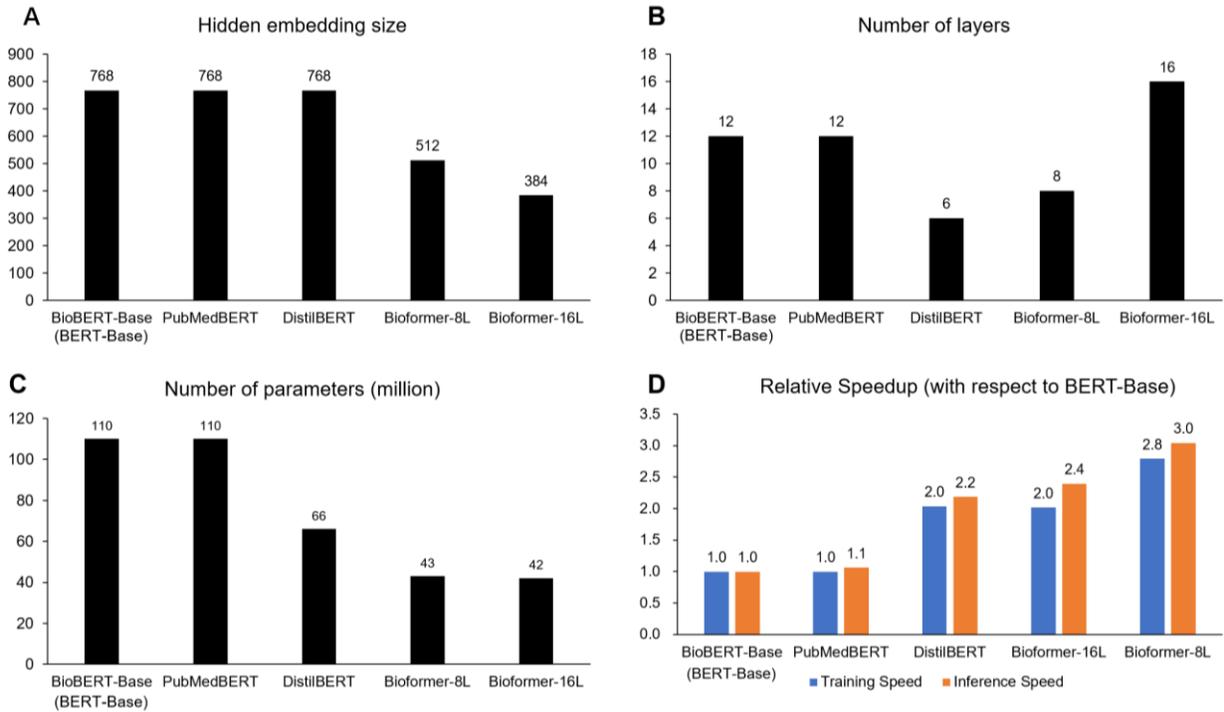

**Figure 3**. **Comparison of hyperparameters and speed.** Hidden embedding sizes, numbers of layers and numbers of parameters of five BERT models are shown in (**A**), (**B**) and (**C**). Relative speedup (with respect to $BioBERT_{Base}/BERT_{Base}$) is shown in (**D**). Training and inference speed are calculated based on a sequence classification task which adds a single linear layer on top of the [CLS] token of the BERT model. Training speed was assessed on an NVIDIA Tesla P100 GPU. Inference speed was assessed on an Intel Xeon CPU.

# Tables

**Table 1. Comparison of Bioformer and other biomedical BERT models.**

|  | Bioformer$_{8L}$ | Bioformer$_{16L}$ | BioBERT$_{Base-v1.1}$ | PubMedBERT |
|---|---|---|---|---|
| Number of parameters | 43M | 42M | 110M | 110M |
| Number of layers | 8 | 16 | 12 | 12 |
| Hidden embedding size | 512 | 384 | 768 | 768 |
| Number of attention heads | 8 | 6 | 12 | 12 |
| Vocabulary size | 32,768 | 32,768 | 28,996 | 30,522 |
| Max input sequence length | 512 | 1,024 | 512 | 512 |
| Pretraining strategy | From scratch | From scratch | Continue pretraining | From scratch |
| Batch size | 256 | 256 | 192 | 8,192 |
| Pre-trained steps | 2M | 2M | 1M | 0.62M |
| Pretraining device | 1 TPUv2 (8 cores) | 1 TPUv2 (8 cores) | 8 V100 GPUs | 16 V100 GPUs |

Table 2. Statistics of the benchmark datasets.

| Dataset | Train | Valid | Test | Metric |
|---|---|---|---|---|
| **Named entity recognition** | | | | |
| BC2GM[32] | 15197 | 3061 | 6325 | F1 score |
| BC4CHEMD[33] | 29478 | 29486 | 25346 | F1 score |
| BC5CDR-chem[34] | 5,203 | 5,347 | 5,385 | F1 score |
| BC5CDR-disease[34] | 4,182 | 4,244 | 4,424 | F1 score |
| JNLPBA[35] | 32178 | 8575 | 6241 | F1 score |
| linnaeus[36] | 2119 | 711 | 1433 | F1 score |
| NCBI-disease[37] | 5134 | 787 | 960 | F1 score |
| s800[38] | 2557 | 384 | 767 | F1 score |
| **Relation extraction** | | | | |
| Chemprot[39] | 4,154 | 2,416 | 3,458 | macro-F1 |
| DDI[40] | 2,937 | 1,004 | 979 | micro F1 |
| euadr[41] | 4796 | 0 | 535 | F1 score |
| GAD-10[42] | 318 | 0 | 38 | F1 score |
| **Document classification** | | | | |
| HoC[43] | 1,108 | 157 | 315 | micro F1 |
| BioCreative-LitCovid[29] | 24,960 | 6,239 | 2,500 | micro F1 |

**Table 3. Performance on downstream biomedical NLP tasks**

| Dataset (metric) | *Bioformer$_{8L}$* | *Bioformer$_{16L}$* | *BioBERT$_{Base-v1.1}$* | *PubMedBERT$_{Abs}$* | *PubMedBERT$_{AbsFull}$* |
|---|---|---|---|---|---|
| **Named entity recognition (NER)** | | | | | |
| BC2GM (F1) | 83.95 | 84.26 | 84.07 | 84.66 | 84.71 |
| BC4CHEMD (F1) | 92.04 | 92.34 | 92.02 | 92.49 | 92.63 |
| BC5CDR-chem (F1) | 93.64 | 94.00 | 93.67 | 94.20 | 94.18 |
| BC5CDR-disease (F1) | 86.29 | 86.53 | 86.09 | 87.54 | 87.17 |
| JNLPBA (F1) | 76.75 | 77.17 | 76.81 | 77.14 | 77.34 |
| Linnaeus (F1) | 88.48 | 88.51 | 84.33 | 88.59 | 88.45 |
| NCBI-disease (F1) | 87.99 | 87.73 | 88.48 | 88.52 | 87.09 |
| S800 (F1) | 73.47 | 74.65 | 74.78 | 74.02 | 74.21 |
| **NER average** | **85.32** | **85.65** | **85.03** | **85.90** | **85.72** |
| **Relation extraction (RE)** | | | | | |
| Chemprot (macro F1) | 76.77 | 79.07 | 77.69 | 80.10 | 80.25 |
| DDI (micro F1) | 81.84 | 84.56 | 82.84 | 84.54 | 83.74 |
| EU-ADR (F1) | 84.72 | 84.76 | 84.21 | 83.60 | 83.88 |
| GAD-10 (F1) | 81.31 | 81.40 | 81.94 | 82.65 | 82.65 |
| **RE average** | **81.16** | **82.45** | **81.67** | **82.72** | **82.63** |
| **Document classification (DC)** | | | | | |
| HoC (micro F1) | 86.41 | 86.37 | 86.42 | 86.41 | 87.07 |
| BC7-LitCovid (micro F1) | 90.88 | 90.69 | 90.72 | 90.78 | 90.78 |
| **DC average** | **88.65** | **88.53** | **88.57** | **88.60** | **88.93** |
| **Question Answering (pretrained on SQuAD 1.1)** | | | | | |
| BioASQ-7b (S. Accuracy) | 40.87 | 43.09 | 42.96 | 43.01 | 42.98 |
| BioASQ-7b (L. Accuracy) | 59.75 | 60.49 | 58.64 | 58.66 | 58.65 |
| BioASQ-7b (MRR) | 48.08 | 49.77 | 49.28 | 49.29 | 49.30 |
| **Question Answering (without pretraining)** | | | | | |
| BioASQ-7b (S. Accuracy) | 34.94 | 37.53 | 35.68 | 35.72 | 35.71 |
| BioASQ-7b (L. Accuracy) | 53.83 | 56.42 | 49.88 | 49.91 | 49.89 |
| BioASQ-7b (MRR) | 41.61 | 44.38 | 41.09 | 41.13 | 41.11 |
| **QA average** | **46.51** | **48.61** | **46.26** | **46.29** | **46.27** |
| **Overall average** | **82.07** | **82.71** | **82.02** | **82.77** | **82.69** |

Notes: 1) BC7-LitCovid denotes BioCreative VII LitCovid challenge and the results are development set performance. 2) S. Accuracy: strict accuracy; L. Accuracy: lenient accuracy; MRR: mean reciprocal rank. 3) QA average is the average score of the three metrics of the two methods (with/without pretraining on SQuAD 1.1). 4) Overall average = (NER average * 8 + RE average * 4 + DC average * 2 + QA average)/15.